\documentclass[letterpaper]{article} % DO NOT CHANGE THIS
\usepackage{aaai2027}  % DO NOT CHANGE THIS
\usepackage[hyphens]{url}  % DO NOT CHANGE THIS
\usepackage{graphicx} % DO NOT CHANGE THIS
\usepackage{natbib}  % DO NOT CHANGE THIS AND DO NOT ADD ANY OPTIONS TO IT
\usepackage{caption} % DO NOT CHANGE THIS AND DO NOT ADD ANY OPTIONS TO IT
\usepackage{algorithm}
\usepackage{algorithmic}

\usepackage{newfloat}
\usepackage{listings}
\DeclareCaptionStyle{ruled}{labelfont=normalfont,labelsep=colon,strut=off} % DO NOT CHANGE THIS
\floatstyle{ruled}
\newfloat{listing}{tb}{lst}{}
\floatname{listing}{Listing}

\usepackage{booktabs}
\usepackage{colortbl}
\usepackage{multirow}
\usepackage{tabularx}
\usepackage{pifont}
\newcommand{\cmark}{\ding{51}}
\newcommand{\xmark}{\ding{55}}
\newcolumntype{P}{>{\centering\arraybackslash}p{1.7em}}
\definecolor{mecoorange}{RGB}{248,232,224}
\definecolor{mecoredorange}{RGB}{255,226,214}

\title{Learning 4D Geometric Priors for Inference-Efficient World Action Models}
\author{
Jianjun Zhang\textsuperscript{1,2,*} \quad
Jian Zhu\textsuperscript{2,\ddag} \quad
Taiyi Su\textsuperscript{2} \quad
Chong Ma\textsuperscript{1,2} \quad
Zitai Huang\textsuperscript{1,2} \\
Yi Xu\textsuperscript{2, \dag} \quad
Hanli Wang\textsuperscript{1,\dag}
}

\affiliations{
\textsuperscript{1}Tongji University \quad
\textsuperscript{2}AIRC, Midea Group \\
\textsuperscript{\dag}Corresponding Author \quad
\textsuperscript{\ddag}Project Leader \\
Project Page: \url{meco-wam.github.io}
}

\nocopyright

\begin{document}

\maketitle

\begingroup
\renewcommand\thefootnote{\textsuperscript{*}}
\footnotetext{This work was completed during an internship at Midea AI Research Centers.}
\endgroup

\begin{abstract}
World Action Models (WAMs) have shown strong potential for robotic manipulation by jointly modeling visual future dynamics and executable action sequences. However, existing video-action co-training methods primarily optimize appearance-oriented video latents, which may insufficiently capture the temporally evolving geometry required for precise manipulation. We propose \textbf{MECo-WAM}, a Multi-Expert Co-Training World Action Model that injects action-relevant 4D geometric priors into video-action representations while preserving the original lightweight inference graph. During training, MECo-WAM combines video and action experts with a lightweight 4D expert supervised by relational targets from a frozen VGGT encoder. Asymmetric expert visibility prevents non-causal shortcuts from auxiliary geometry to action generation. To transfer geometric knowledge into the deployed video-action pathway, we introduce \emph{decayed 4D read-mask attention}, which provides restricted current-frame geometric guidance early in training and progressively removes this dependency. We further propose \emph{action-aware temporal geometric distillation}, which aligns within-frame geometric relations and their temporal evolution while emphasizing visual regions most relevant to robot actions. At deployment, all auxiliary 4D components are removed. Experiments on LIBERO (98.2\%), RoboTwin 2.0 (92.6\%), and challenging real-world manipulation tasks show that MECo-WAM improves manipulation performance without increasing inference cost.
\end{abstract}

\begin{figure}[!t]
\centering
\includegraphics[width=1.00\columnwidth]{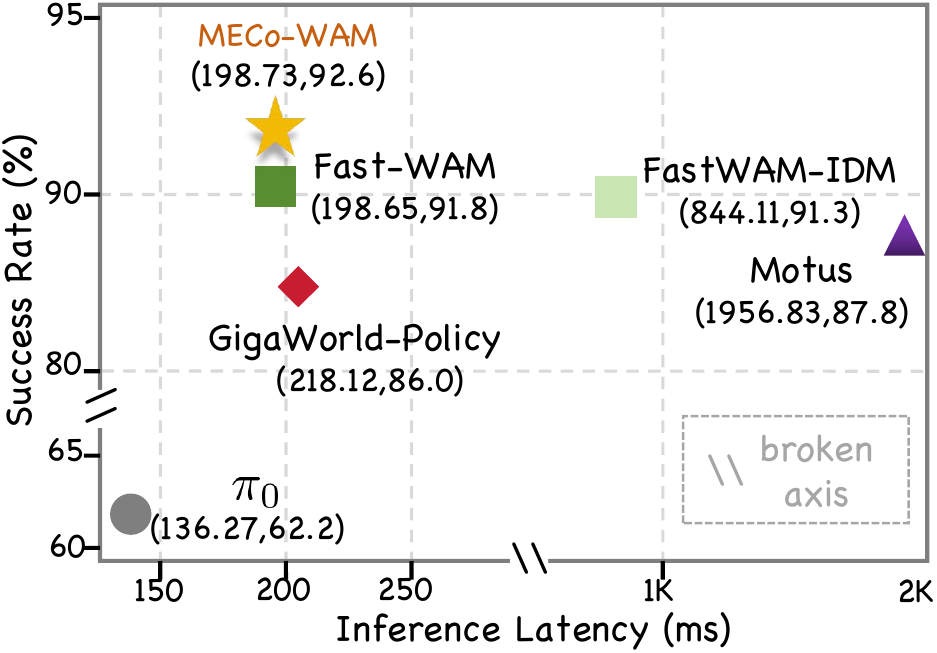}
\caption{Comparison of MECo-WAM with Baselines in action-chunk inference latency and task success rate on RoboTwin.}
\label{fig:robotwin_sr_latency}
\end{figure}

\section{Introduction}

Robotic manipulation requires a policy to map visual observations and language instructions to precise action trajectories \cite{hu25vpp,ma2026dit4dit,su2026demavla,wang2026wam,su2025mtv}. World action models offer a promising formulation by jointly learning how visual states evolve under interaction and how robot actions should be generated \cite{ye2026dreamzero,kim2026cosmospolicy,bi2026motus,li2026lingbotva}. Compared with direct action policies, video-action co-training can provide richer motion and interaction priors, enabling the policy to reason over changes that unfold beyond a single observation \cite{ye2026gigaworld,yuan2026fastwam}.

Despite this advantage, the visual representation learned by many WAMs remains dominated by appearance-oriented video prediction. A video latent can support plausible future synthesis without explicitly preserving the spatial relations that determine whether a grasp is reachable, whether an object is aligned with a target, or whether contact will cause a stable transition. These relations are not static: they evolve as the robot approaches, contacts, moves, and releases objects. Recent geometry-aware VLA and WAM studies therefore introduce 3D or 4D structure to strengthen spatial grounding for manipulation \cite{qu2025spatialvla,li2025bridgevla,li20253dsvla,guo2026xwam,li2026wam4d}. Thus, manipulation-oriented WAMs require geometry-aware temporal representations beyond visual plausibility.

A direct approach is to introduce explicit 4D reconstruction or dense geometric prediction into the world action modeling pipeline \cite{guo2026xwam,li2026wam4d}. However, making geometry an explicit deployment-time output increases inference cost and may shift optimization toward geometric reconstruction that is only weakly coupled with action generation. More importantly, generic geometric supervision does not distinguish the visual relations that are causally relevant to the robot's current action. More importantly, generic geometric supervision may overlook action-relevant relations among manipulated objects, target regions, and their temporal interactions.

We therefore ask a focused question: \emph{can a world action model acquire action-relevant temporal geometry during training while retaining the same lightweight video-action inference graph at deployment?} Our answer is \textbf{MECo-WAM}, a Multi-Expert Co-Training World Action Model. MECo-WAM adds a lightweight 4D expert only during training, where frozen VGGT features \cite{wang2025vggt} supervise temporal geometric prediction alongside video and action denoising. Rather than allowing unrestricted cross-expert communication, the deployed video-action pathway receives only restricted current-frame geometric guidance during early training, while future geometry remains loss-side supervision. This design transfers 4D priors without creating non-causal shortcuts or adding inference-time cost. As shown in Figure~\ref{fig:robotwin_sr_latency}, this design achieves a strong task success rate on RoboTwin 2.0 \cite{mu2025robotwin} while keeping action-chunk inference latency low.

To transfer geometry without a permanent 4D dependency, we introduce \emph{decayed 4D read-mask attention}, which exposes only the current-frame geometry token early in training and removes this access before deployment. We further propose \emph{action-aware temporal geometric distillation}, aligning predicted 4D keyframes and their temporal relation changes with frozen VGGT targets while emphasizing action-relevant token pairs.

Our contributions are fourfold:
\begin{itemize}
\item We propose \textbf{MECo-WAM}, a multi-expert co-training framework that injects action-aware 4D geometric priors into WAM representations while preserving the original lightweight inference graph.

\item We introduce \textbf{decayed 4D read-mask attention}, which provides early-stage geometric guidance and progressively removes the dependency on 4D tokens before deployment.

\item We propose \textbf{action-aware temporal geometric distillation}, which aligns within-frame geometric relations and their temporal evolution while emphasizing action-relevant visual regions, enabling task-conditioned geometry learning.

\item Extensive experiments on LIBERO, RoboTwin 2.0, and real-world manipulation tasks demonstrate consistent gains in task success and execution efficiency.
\end{itemize}

\begin{figure*}[t]
\centering
\includegraphics[width=\textwidth]{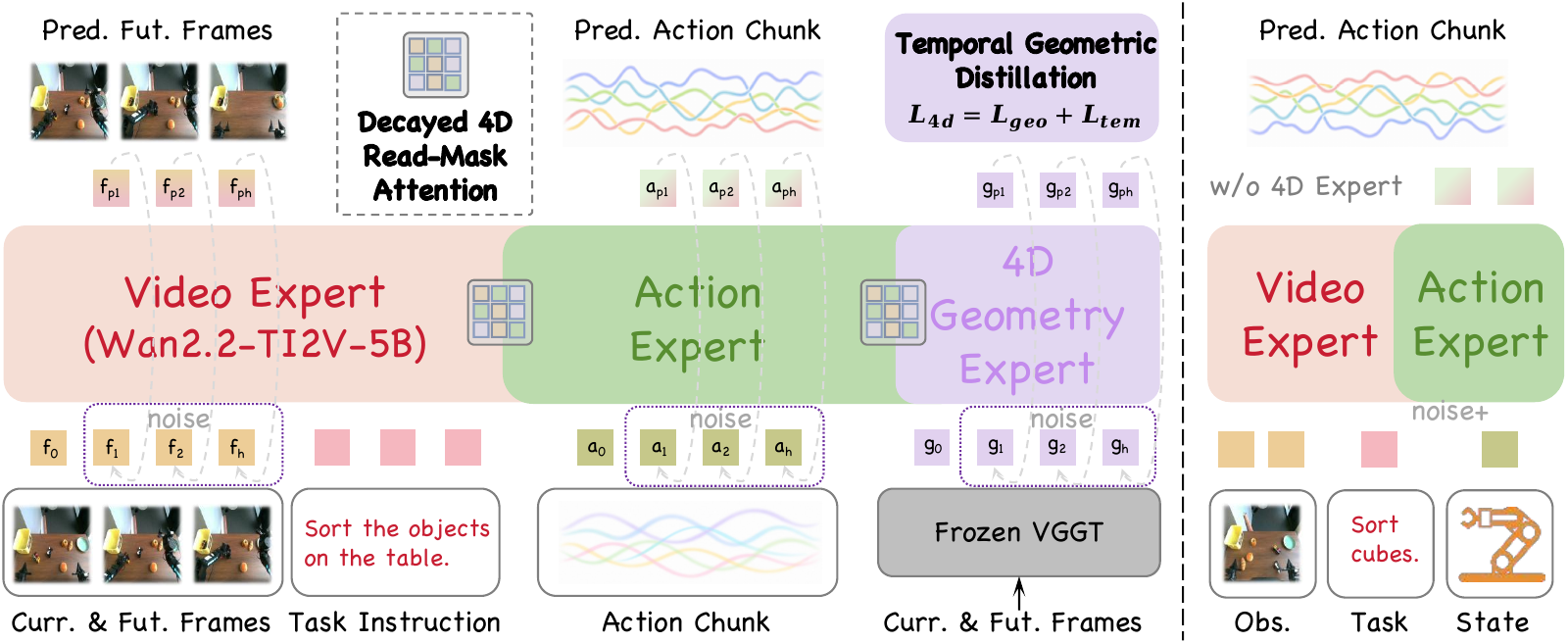}
\caption{Overview of MECo-WAM. The left side illustrates the training process, while the right side shows the inference process. During training, video frames are encoded by the VAE for video denoising, while current and future frames are encoded by a frozen VGGT encoder to provide targets for 4D geometry denoising. The 4D expert takes $g_0$, $g_1$, $g_2$, and $g_h$ as input slots, predicts $g_{p1}$, $g_{p2}$, and $g_{ph}$, and applies keyframe 4D losses on selected predictions. Decayed 4D read-mask attention transfers early current-frame geometric guidance to the video-action pathway. At inference, only the original video and action experts remain.}
\label{fig:overview}
\end{figure*}

\section{Related Work}

\subsection{World Action Models}

World action models bridge two complementary embodied learning paradigms: policies that map observations and instructions to executable actions, and world models that predict how the environment evolves under interaction. Direct VLA policies, including the $\pi$ model family \cite{black2024pi0,intelligence2025pi05} and OpenVLA \cite{kim2025openvla}, provide strong reactive observation-to-action baselines. Recent WAMs adapt pretrained video priors or unified multimodal architectures to jointly model future observations, latent dynamics, and action chunks. DreamZero \cite{ye2026dreamzero}, Mimic-Video \cite{pai2025mimic}, Cosmos Policy \cite{kim2026cosmospolicy}, Motus \cite{bi2026motus}, and LingBot-VA \cite{li2026lingbotva} show that video-based temporal supervision can improve policy learning beyond direct behavior cloning by exposing the model to physical evolution and action-conditioned scene changes.

Some systems improve practical execution through causal attention, multi-chunk prediction, caching, asynchronous denoising, or action-centered interfaces \cite{ye2026gigaworld,xu2026nextforcing,team2026motubrain,guo2026xwam}. Fast-WAM \cite{yuan2026fastwam} makes this deployment-oriented view explicit: future video prediction remains useful during training, while test-time action generation can proceed without explicit future imagination. MECo-WAM follows this efficient WAM principle but studies a different source of supervision. Rather than adding rollout to the deployed policy, it uses a training-only 4D expert to transfer action-relevant temporal geometry into the shared video-action representation. The inference graph therefore remains the same lightweight observation-to-action path used by the base WAM.

\subsection{Geometry-Aware Embodied Models}

Geometry-aware embodied models improve manipulation by injecting 3D structure through depth, point clouds, spatial priors, or representation alignment. In VLA models, 3D-VLA connects 3D perception, reasoning, and action through a generative world model \cite{zhen20243dvla}. SpatialVLA \cite{qu2025spatialvla} studies spatial representations for action prediction, BridgeVLA \cite{li2025bridgevla} aligns 3D inputs and heatmap-style outputs in a shared 2D space, 3DS-VLA \cite{li20253dsvla} introduces 3D spatial constraints for robust multi-task manipulation, and GeoVLA \cite{sun2025geovla} strengthens VLA policies with explicit 3D representations. Recent variants \cite{fan2026any3dvla,ni2026swiftvla,spiridonov2025generalist,zhang2026falcon,qian2026geopredict,ye2026st4vla} further improve robustness or data efficiency through diverse point clouds, lightweight spatiotemporal dynamics, manipulation data beyond action labels, spatial foundation priors, predictive kinematics with 3D Gaussian geometry, and spatially guided training. Spatial Forcing  \cite{li2026spatialforcing} further shows that spatial foundation priors can be transferred into VLA representations through implicit spatial alignment.

In the WAM setting, geometry has begun to move from direct policy inputs into future prediction and video-action co-training. X-WAM \cite{guo2026xwam} unifies action execution with 4D world synthesis by predicting multi-view RGB-D futures, adding a lightweight depth branch to a pretrained video diffusion backbone, and using asynchronous denoising for efficient action decoding. WAM4D \cite{li2026wam4d} studies fast 4D world action modeling with spatial register tokens and future-depth readouts, then removes the register branch for action inference. MECo-WAM follows this geometry-for-WAM direction but differs in where geometry lives: it treats frozen VGGT \cite{wang2025vggt} 4D structure as a training-time representation constraint, uses decayed read-mask attention to avoid permanent dependence, and transfers temporal geometry into the lightweight video-action path rather than requiring explicit 4D reconstruction during deployment.

\section{Methodology}

\subsection{Problem Formulation}

Figure~\ref{fig:overview} gives an overview of MECo-WAM. Let $o_0$ denote the current visual observation, $a_0$ the current robot-state/proprioceptive context, $\ell$ a language instruction, and $a_{1:H}$ an action chunk of horizon $H$. The deployed policy models
\begin{equation}
p_{\theta}(a_{1:H}\mid o_0,a_0,\ell).
\label{eq:policy}
\end{equation}
During training, the model observes visual trajectories and action chunks, and the base video-action pathway learns future visual dynamics together with action denoising. MECo-WAM adds a lightweight 4D expert only for training, using frozen VGGT \cite{wang2025vggt} features from current and future frames as geometric supervision. This auxiliary path constrains the shared representation through relational 4D losses, while the deployed policy remains an observation-to-action model.

Our design principle is that \emph{4D geometry serves as a training-time representation constraint rather than an inference-time input or output.} At deployment, the 4D expert, frozen VGGT encoder, and alignment modules are removed, preserving the original observation-to-action interface.

\subsection{Multi-Expert Co-Training Architecture}

\paragraph{Expert tokens.}
The video expert uses the clean first-frame VAE token as visual context and denoises noisy future VAE target slots. Let $\bar{Y}_v=[\bar{f}_1,\bar{f}_2,\bar{f}_h]$ denote the clean future VAE targets. For flow-matching time $r$, we construct noisy future slots
\begin{equation}
\begin{array}{c}
[f_1,f_2,f_h]=(1-r)\bar{Y}_v+r\epsilon_v,\\[4pt]
X_v=[f_0,f_1,f_2,f_h],\\[4pt]
\left[f_{p1},f_{p2},f_{ph}\right]=E_v(X_v,r;\ell),
\end{array}
\label{eq:video_tokens}
\end{equation}
where $f_{p1}$, $f_{p2}$, and $f_{ph}$ are the corresponding predicted future video outputs supervised against $\bar{Y}_v$ through the video loss. The action expert uses the same noise-slot convention, with $a_0$ serving as a clean robot-state/proprioceptive anchor rather than a prediction target. Let $\bar{Y}_a=[\bar{a}_1,\bar{a}_2,\bar{a}_h]$ denote clean future action targets:
\begin{equation}
\begin{array}{c}
[a_1,a_2,a_h]=(1-r)\bar{Y}_a+r\epsilon_a,\\[4pt]
X_a=[a_0,a_1,a_2,a_h],\\[4pt]
\left[a_{p1},a_{p2},a_{ph}\right]=E_a(X_a,r;f_0,\ell).
\end{array}
\label{eq:action_tokens}
\end{equation}
The 4D expert uses the same denoising convention, but obtains its tokens from a frozen VGGT encoder \cite{wang2025vggt} instead of the VAE. Given current and future RGB frames, frozen VGGT produces clean geometry targets $g_0$ and $\bar{Y}_g=[\bar{g}_1,\bar{g}_2,\bar{g}_h]$. The current geometry token remains clean, while future geometry slots are noisy:
\begin{equation}
\begin{array}{c}
[g_1,g_2,g_h]=(1-r)\bar{Y}_g+r\epsilon_g,\\[4pt]
X_{4\mathrm{d}}=[g_0,g_1,g_2,g_h],\\[4pt]
\left[g_{p1},g_{p2},g_{ph}\right]=E_{4\mathrm{d}}(X_{4\mathrm{d}},r).
\end{array}
\label{eq:4dpred}
\end{equation}
The 4D objective is applied only on selected keyframe predictions. Let $\mathcal{K}\subseteq\{1,2,h\}$ denote the selected keyframe indices:
\begin{equation}
\widehat{G}_{\mathcal{K}}=\{g_{pk}\mid k\in\mathcal{K}\},
\qquad
G_{\mathcal{K}}=\{\bar{g}_k\mid k\in\mathcal{K}\}.
\label{eq:4dselect}
\end{equation}
Clean VGGT targets $G_{\mathcal{K}}$ are used only for training losses and never become input to the deployed policy.

The mixed-attention sequence is
\begin{equation}
X=[X_v,X_a,X_{4\mathrm{d}}].
\label{eq:tokenorder}
\end{equation}
At each transformer layer, expert $e\in\{v,a,4\mathrm{d}\}$ has separate query, key, and value projections:
\begin{equation}
Q_e=X_eW_e^Q,\qquad K_e=X_eW_e^K,\qquad V_e=X_eW_e^V.
\label{eq:qkv}
\end{equation}
After concatenating the expert-specific tensors, masked mixed attention produces
\begin{equation}
Y=\mathrm{softmax}\left(\frac{QK^{\top}}{\sqrt{d}}+M\right)V,
\label{eq:attention}
\end{equation}
where $M$ is an expert-level attention mask. The mask defines the information paths that are allowed during co-training.

\subsection{Decayed 4D Read-Mask Attention}

MECo-WAM introduces decayed 4D read-mask attention to preserve video-geometry co-training benefits while decoupling the 4D path at inference. The mask prevents future-information shortcuts: current anchors $f_0$, $a_0$, and $g_0$ are self-only; future video tokens read the video branch, but not action tokens; and future action tokens read clean visual context plus the action branch, but not noisy future video slots. MECo-WAM then adds temporary read edges from future video/action queries to the current-frame geometry token $g_0$, which is safe because it is encoded from the current RGB frame only. Future 4D slots stay inside the 4D branch and serve only as auxiliary prediction targets.

Figure~\ref{fig:mask_attention} visualizes the proposed visibility mask and highlights the decayed 4D attention edges used only during training.
The read edges are stochastic and decay over optimization. Let $s$ be the optimization step and let $\gamma_s$ indicate whether the $g_0$ read edges are active:

\begin{figure}[t!]
\centering
\includegraphics[width=\columnwidth]{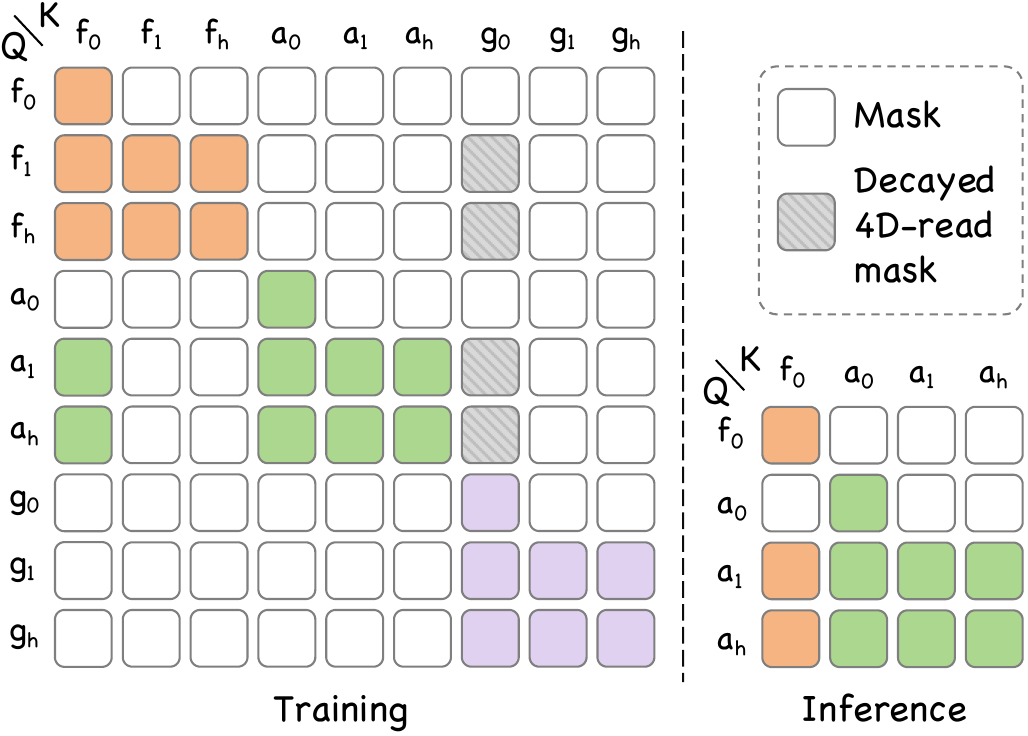}
\caption{Decayed read-mask attention. Rows and columns denote query and key slots. Hatched cells are temporary reads from future video/action queries to current geometry $g_0$; white cells are masked. Future 4D slots are auxiliary training targets, and all 4D tokens are removed at inference.}
\label{fig:mask_attention}
\end{figure}

\begin{equation}
\gamma_s\sim\mathrm{Bernoulli}\bigl(p_{4\mathrm{d}}(s)\bigr).
\label{eq:readsample}
\end{equation}
The activation probability follows a linear decay:
\begin{equation}
p_{4\mathrm{d}}(s)=
\left\{
\begin{array}{ll}
p_{\mathrm{start}}+
\left(p_{\mathrm{end}}-p_{\mathrm{start}}\right)
\frac{s}{S_{\mathrm{decay}}}, & s<S_{\mathrm{decay}},\\[4pt]
p_{\mathrm{end}}, & s\geq S_{\mathrm{decay}}.
\end{array}
\right.
\label{eq:readprob}
\end{equation}
The read edges are present only when $\gamma_s=1$. This schedule exposes the video-action path to current-frame geometry early in training and gradually removes the auxiliary dependency. At deployment, $p_{\mathrm{end}}=0$, no 4D tokens are instantiated, and the original video-action graph is recovered.

\subsection{Action-Aware Temporal Geometric Distillation}

The distillation objective uses frozen VGGT as a geometry teacher, but supervises only the predictions of the training-time 4D expert. Its mechanism has three parts. First, relation matching transfers the 3D layout without requiring the student and teacher to share the same absolute feature coordinates. Second, action-aware weights identify which visual tokens are most coupled with the contemporaneous robot motion, so the loss focuses on manipulated objects, targets, and contact regions. Third, temporal relation matching teaches how these action-relevant relations change across keyframes, capturing approach, contact, transport, and release dynamics rather than only static scene geometry.

\begin{figure}[t]
\centering
\includegraphics[width=\columnwidth]{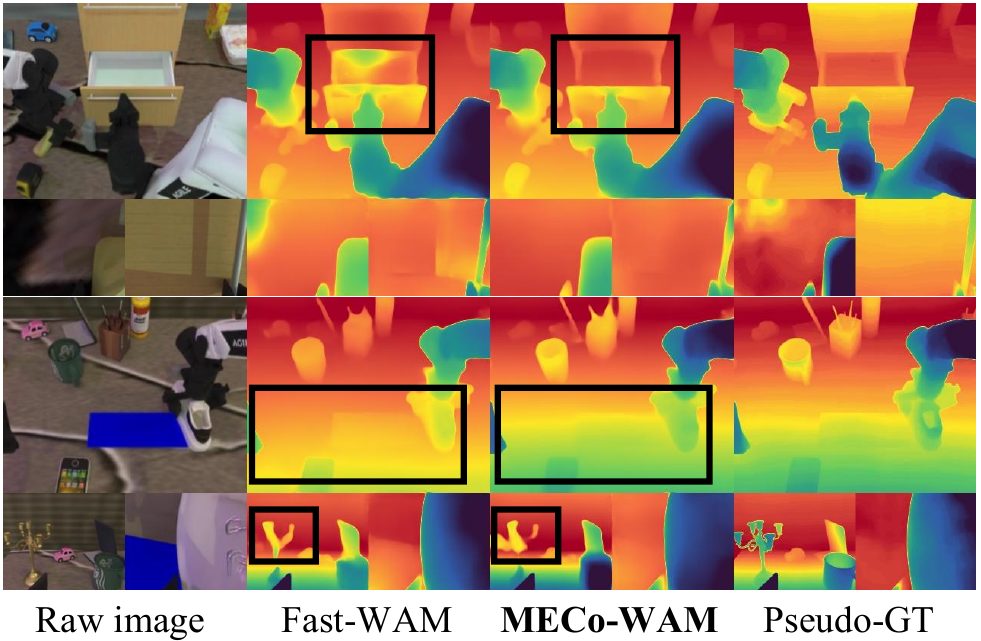}
\caption{Depth probing of shared video-action representations. A matched DPT-style head is trained on tokens from the final four layers of Fast-WAM or MECo-WAM. The improved MECo-WAM depth structure indicates that 4D co-training enriches the deployed video-action representation. Pseudo-GT depth maps are generated by VDA \cite{chen2025vda}.}
\label{fig:depthprobe}
\end{figure}

\paragraph{VGGT-aligned geometry.}
For selected keyframes $k\in\mathcal{K}$, the student geometry features are
\begin{equation}
Z^k=P_{\mathrm{align}}(g_{pk}),\qquad k\in\mathcal{K},
\label{eq:student}
\end{equation}
and the frozen VGGT encoder provides the corresponding clean target geometry $G_T^k=\bar{g}_k$. We represent geometry by pairwise feature relations:
\begin{equation}
\begin{array}{c}
R_{4\mathrm{d}}^k(i,j)=\left\|Z_i^k-Z_j^k\right\|_2,\\[4pt]
R_T^k(i,j)=\left\|G_{T,i}^k-G_{T,j}^k\right\|_2.
\end{array}
\label{eq:relations}
\end{equation}
We normalize the valid entries of each relation matrix before alignment, denoting the normalized relations by $\widehat{R}_{4\mathrm{d}}^k$ and $\widehat{R}_T^k$. Relational matching avoids assuming a shared absolute feature coordinate system and focuses supervision on relative object-gripper-target structure.

\paragraph{Action-aware weights.}
Not all visual regions contribute equally to a given action. Let $v_i^k$ be a selected video token and $\bar{a}_k$ a pooled action representation aligned to the same temporal segment. We compute
\begin{equation}
s_i^k=
\frac{(W_v v_i^k)^{\top}(W_a\bar{a}_k)}
{\tau\left\|W_v v_i^k\right\|_2\left\|W_a\bar{a}_k\right\|_2},
\qquad
r_i^k=\mathrm{softmax}_i(s_i^k).
\label{eq:relevance}
\end{equation}
The final token weights mix the learned relevance with a uniform prior:
\begin{equation}
w_i^k=N\left((1-\eta)r_i^k+\frac{\eta}{N}\right),
\qquad
w_{ij}^k=\sqrt{w_i^kw_j^k},
\label{eq:weights}
\end{equation}
where $N$ is the number of selected tokens and $\eta$ prevents collapse onto a single location. The within-frame loss is
\begin{equation}
\mathcal{L}_{\mathrm{geo}}^{\mathrm{act}}=
\frac{\sum_{k\in\mathcal{K},i,j}w_{ij}^k
\left|\widehat{R}_{4\mathrm{d}}^k(i,j)-\widehat{R}_T^k(i,j)\right|}
{\sum_{k\in\mathcal{K},i,j}w_{ij}^k}.
\label{eq:geoloss}
\end{equation}
This weighting makes the 4D loss action-conditioned. High-relevance tokens contribute more strongly to pairwise geometry, while the uniform mixture prevents the loss from collapsing onto a single region and preserves useful scene context.

\paragraph{Temporal geometry.}
Manipulation also depends on how geometry changes. For consecutive selected keyframe pairs $(k,k^+)\in\mathcal{A}_{\mathcal{K}}$, we define normalized relation change:
\begin{equation}
\Delta(R_k,R_{k^+})=
\frac{R_{k^+}-R_k}
{\left|R_{k^+}\right|+\left|R_k\right|+\epsilon}.
\label{eq:delta}
\end{equation}
Adjacent-frame pair weights are defined as
\begin{equation}
w_{ij}^{k,k^+}=\sqrt{w_{ij}^kw_{ij}^{k^+}}.
\label{eq:temporalweights}
\end{equation}
For compactness, define
\begin{equation}
D_{ij}^{k,k^+}=
\bigl|
\Delta(\widehat{R}_{4\mathrm{d}}^k,\widehat{R}_{4\mathrm{d}}^{k^+})
-
\Delta(\widehat{R}_T^k,\widehat{R}_T^{k^+})
\bigr|.
\label{eq:temporal_distance}
\end{equation}
The temporal loss is
\begin{equation}
\mathcal{L}_{\mathrm{tem}}^{\mathrm{act}}=
\frac{\sum_{(k,k^+)\in\mathcal{A}_{\mathcal{K}},i,j}w_{ij}^{k,k^+}
D_{ij}^{k,k^+}}
{\sum_{(k,k^+)\in\mathcal{A}_{\mathcal{K}},i,j}w_{ij}^{k,k^+}}.
\label{eq:temloss}
\end{equation}
This temporal objective complements static relation matching by supervising how action-relevant geometry evolves between keyframes.

\subsection{Training Objective and Inference}

The video and action experts follow the conditional flow-matching objective of the base WAM. For target $y$, Gaussian noise $\epsilon$, and interpolation time $r$, we construct
\begin{equation}
y_r=(1-r)y+r\epsilon,
\label{eq:interpolation}
\end{equation}
and optimize
\begin{equation}
\mathcal{L}_{\mathrm{FM}}(y)=
\mathrm{E}_{y,\epsilon,r}
\left[
\left\|
u_{\theta}(y_r,r,o_0,a_0,\ell)-(\epsilon-y)
\right\|_2^2
\right].
\label{eq:fm}
\end{equation}
Instantiating this loss for future video latents and action chunks yields $\mathcal{L}_{\mathrm{video}}$ and $\mathcal{L}_{\mathrm{action}}$. The 4D objective is
\begin{equation}
\mathcal{L}_{4\mathrm{d}}=
\alpha_{\mathrm{geo}}\mathcal{L}_{\mathrm{geo}}^{\mathrm{act}}+
\alpha_{\mathrm{tem}}\mathcal{L}_{\mathrm{tem}}^{\mathrm{act}},
\label{eq:4dloss}
\end{equation}
and the total objective is
\begin{equation}
\mathcal{L}_{\mathrm{total}}=
\lambda_{\mathrm{video}}\mathcal{L}_{\mathrm{video}}+
\lambda_{\mathrm{action}}\mathcal{L}_{\mathrm{action}}+
\lambda_{4\mathrm{d}}\mathcal{L}_{4\mathrm{d}}.
\label{eq:total}
\end{equation}

During training, MECo-WAM uses the auxiliary 4D expert, frozen VGGT encoder, and decayed read mask, but these components are removed at inference, leaving only the original video and action experts. Thus, MECo-WAM adds no geometric decoder, sensor input, or extra denoising stage to the deployed policy. Figure~\ref{fig:depthprobe} further suggests that the training-time 4D objective transfers geometric priors into the WAM representation.

\section{Experiments}

\subsection{Experimental Setup}

\paragraph{Implementation details.}
MECo-WAM uses Wan2.2-TI2V-5B \cite{wan2025wan} as the video backbone and follows the Fast-WAM observation-to-action deployment interface \cite{yuan2026fastwam}. The action expert uses 30 DiT blocks, 24 attention heads, 128-dimensional heads, and hidden width $d_a=1024$ (about 1B parameters). The auxiliary 4D expert uses $d_{4\mathrm{d}}=512$ (about 0.45B parameters), with supervision from a frozen VGGT-1B encoder. Decayed 4D read probability decreases linearly from 1.0 to 0 over the first half of training, and final-layer VGGT/MECo-WAM tokens are used for relational alignment.

Each training chunk contains 33 robot steps, corresponding to action horizon $H=32$ and 9 video frames under a $4\times$ action-to-video temporal ratio. We use continuous flow matching with 1000 training timesteps and shift 5.0, AdamW with learning rate $1\times10^{-4}$, weight decay 0.01, cosine decay, bfloat16 mixed precision, and gradient clipping at 1.0. Training is conducted on 64 NVIDIA H20 96GB GPUs. Inference uses 10 denoising steps with CFG scale 1.0 on a single NVIDIA RTX 5090 32GB GPU, and real-world experiments are conducted with an ARX-R5 robotic arm.

\paragraph{Benchmarks.}
We evaluate MECo-WAM on LIBERO \cite{liu2023libero}, RoboTwin 2.0 \cite{mu2025robotwin}, and real-world tabletop manipulation. For simulation experiments, we follow the Fast-WAM evaluation configuration \cite{yuan2026fastwam}. LIBERO covers four suites: Spatial, Object, Goal, and Long, each containing 10 tasks with 500 expert demonstrations. RoboTwin 2.0 evaluates bimanual manipulation under clean and randomized conditions, where randomization changes object poses, appearances, clutter, illumination, and tabletop layouts. For real-world evaluation, we use two tabletop tasks, stacking three sponge cubes vertically and sorting three cubes into a size-ordered line, and report success rate, progress rate, correction count, and completion time under an identical camera setup and execution budget.

\begin{table}[t]
\centering
\setlength{\tabcolsep}{2.8pt}
\renewcommand{\arraystretch}{1.0}
\begin{tabular}{lccccc}
\toprule
\multirow{2}{*}{\textbf{Method}} & \multicolumn{5}{c@{}}{\textbf{LIBERO}} \\
\cmidrule(lr){2-6}
& \textbf{Spat.} & \textbf{Obj.} & \textbf{Goal} & \textbf{Long} & \textbf{Avg.} \\
\midrule
\multicolumn{6}{>{\columncolor{black!10}}c}{\textbf{VLA Models}} \\
$\pi_0$                      & 96.8 & 98.8 & 95.8 & 85.2 & 94.1 \\
$\pi_0$+FAST                 & 96.4 & 96.8 & 88.6 & 60.2 & 85.5 \\
OpenVLA                      & 94.4 & 88.4 & 79.2 & 53.7 & 76.5 \\
OpenVLA-OFT                  & 97.6 & 98.4 & \underline{97.9} & 94.5 & 97.1 \\
DD-VLA                       & 97.2 & 96.6 & 97.4 & 92.0 & 96.3 \\
Uni-VLA                      & 95.4 & 98.8 & 93.6 & 94.0 & 95.4 \\
X-VLA                        & 98.2 & 98.6 & 97.8 & \underline{97.6} & 98.1 \\
\midrule
\multicolumn{6}{>{\columncolor{black!10}}c}{\textbf{WAMs}} \\
LingBot-VA (P.T.)            & \underline{98.5} & 99.6 & 97.2 & \textbf{98.5} & \textbf{98.5} \\
Motus (P.T.)                 & 96.8 & \underline{99.8} & 96.6 & \underline{97.6} & 97.7 \\
Fast-WAM (w/o P.T.)          & 98.2 & \textbf{100.0} & 97.0 & 95.2 & 97.6 \\
\rowcolor{mecoorange}\textbf{MECo-WAM (w/o P.T.)} & \textbf{98.8} & \textbf{100.0} & \textbf{98.2} & 95.8 & \underline{98.2} \\
\bottomrule
\end{tabular}
\caption{LIBERO success rates (\%). Spat./Obj./Avg. denote Spatial/Object/Average; P.T. denotes embodied-policy pretraining; bold/underline denote best/second-best.}
\label{tab:libero_main}
\end{table}

\begin{table}[t]
\centering
\setlength{\tabcolsep}{5.2pt}
\begin{tabularx}{\columnwidth}{>{\raggedright\arraybackslash}X|P|c c|c}
\toprule
\textbf{Method} & \textbf{P.T.} & \textbf{Clean} & \textbf{Rand.} & \textbf{Average} \\
\midrule
$\pi_{0.5}$     & \cmark & 82.74 & 76.76 & 79.75 \\
X-VLA           & \cmark & 72.80 & 72.84 & 72.82 \\
Motus           & \cmark & 88.66 & 87.02 & 87.84 \\
LingBot-VA      & \cmark & 92.90 & 91.50 & \underline{92.20} \\
Fast-WAM        & \xmark & 91.88 & 91.78 & 91.83 \\
\rowcolor{mecoorange}\textbf{MECo-WAM} & \xmark & \textbf{93.26} & \textbf{91.98} & \textbf{92.62} \\
\bottomrule
\end{tabularx}
\caption{RoboTwin 2.0 success rates (\%) under clean and randomized evaluation. P.T. denotes embodied-policy pretraining.}
\label{tab:robotwin_main}
\end{table}

\subsection{Main Results}

\paragraph{LIBERO.}
Table~\ref{tab:libero_main} compares MECo-WAM with VLA policies, including $\pi_0$ \citep{black2024pi0}, $\pi_0$+FAST \citep{pertsch2025pifast}, OpenVLA \citep{kim2025openvla}, OpenVLA-OFT \citep{kim2025openvlaoft}, DD-VLA \citep{liang2025ddvla}, Uni-VLA \citep{wang2025univla}, and X-VLA \citep{zheng2025xvla}, as well as WAM-style models, including LingBot-VA \citep{li2026lingbotva}, Motus \citep{bi2026motus}, and Fast-WAM \citep{yuan2026fastwam}. Without embodied-policy pretraining, MECo-WAM reaches 98.2\% average success, improving over Fast-WAM by 0.6 points and Motus by 0.5 points, while remaining close to the pretrained LingBot-VA result (98.5\%). The gains are most apparent on geometry-sensitive suites: MECo-WAM obtains 98.8\% on Spatial, 100.0\% on Object, and 98.2\% on Goal, improving Fast-WAM by 0.6 points on Spatial, matching it on Object, and adding 1.2 points on Goal. These results suggest that action-aware 4D co-training mainly strengthens spatial and object-centric reasoning, which is exactly where WAM policies benefit from better geometric priors.

\paragraph{RoboTwin 2.0.}
Table~\ref{tab:robotwin_main} reports aggregate RoboTwin 2.0 success rates and additionally compares with $\pi_{0.5}$ \citep{intelligence2025pi05}. MECo-WAM obtains the best overall average, improving the non-pretrained Fast-WAM baseline from 91.83\% to 92.62\% while keeping the same inference-time video-action graph. The improvement is larger in the clean setting (93.26\% vs. 91.88\%, +1.38 points) and remains positive under randomization (91.98\% vs. 91.78\%, +0.20 points), indicating that the learned geometric prior helps both nominal execution and visually perturbed scenes. MECo-WAM also slightly surpasses the strongest pretraining-based WAM average in the table, LingBot-VA (92.62\% vs. 92.20\%), showing that training-time 4D supervision can make a non-pretrained WAM competitive without adding geometry modules at deployment.

\begin{figure*}[!t]
\centering
\includegraphics[width=.92\textwidth]{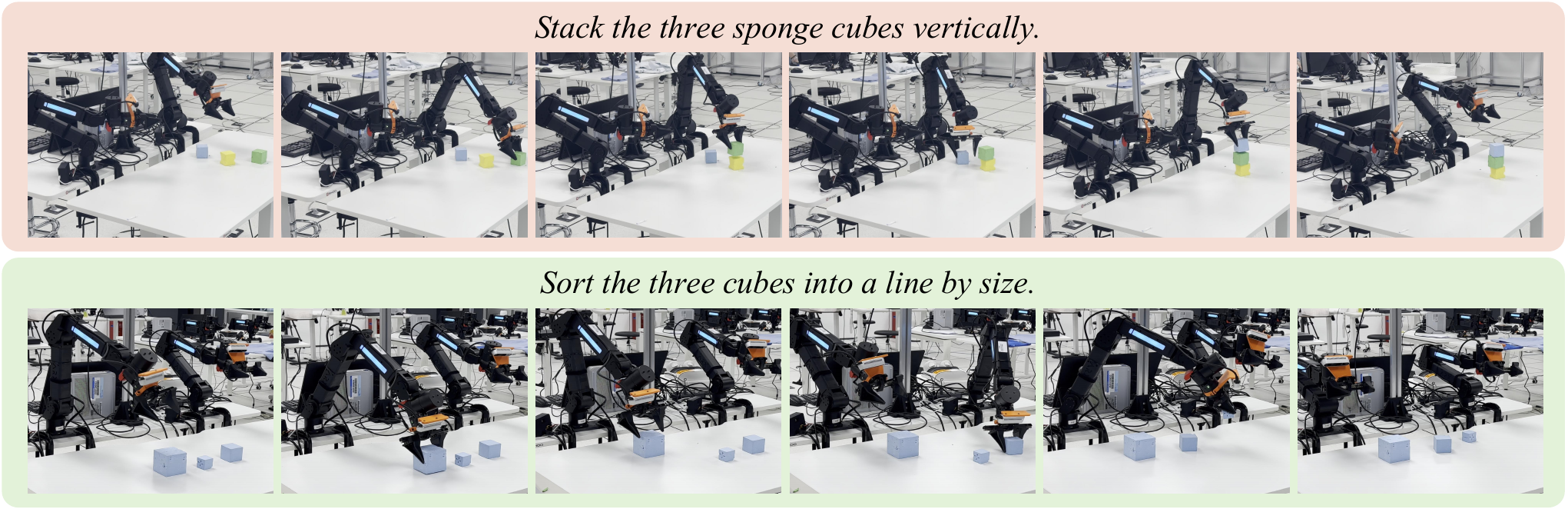}
\caption{Representative real-robot tabletop experiments on cube stacking and size-based cube sorting.}
\label{fig:real_robot}
\end{figure*}

\begin{table*}[!t]
\centering
\scriptsize
\setlength{\tabcolsep}{3pt}
\renewcommand{\arraystretch}{0.9}
\resizebox{\textwidth}{!}{%
\begin{tabular}{l|cccc|cccc|cccc|cccc}
\toprule
\multicolumn{1}{c|}{\multirow{2}{*}{\textbf{Method}}} &
\multicolumn{4}{c|}{\textbf{SR (\%) $\uparrow$}} &
\multicolumn{4}{c|}{\textbf{PR (\%) $\uparrow$}} &
\multicolumn{4}{c|}{\textbf{CR (\#) $\downarrow$}} &
\multicolumn{4}{c}{\textbf{CT (s) $\downarrow$}} \\
\cmidrule(lr){2-5}\cmidrule(lr){6-9}\cmidrule(lr){10-13}\cmidrule(l){14-17}
& \textbf{Test 1} & \textbf{Test 2} & \textbf{Test $n$} & \textbf{Avg}
& \textbf{Test 1} & \textbf{Test 2} & \textbf{Test $n$} & \textbf{Avg}
& \textbf{Test 1} & \textbf{Test 2} & \textbf{Test $n$} & \textbf{Avg}
& \textbf{Test 1} & \textbf{Test 2} & \textbf{Test $n$} & \textbf{Avg} \\
\midrule
\multicolumn{17}{>{\columncolor{black!10}[\tabcolsep][\tabcolsep]}c}{\emph{Stack the three sponge cubes vertically.}} \\
$\pi_0$          & 0 & 100 & 0 & 40.0 & 50 & 100 & 50 & 55.0 & -- & 1 & -- & \textbf{0.75} & -- & 32.62 & -- & 25.98 \\
Fast-WAM         & 100 & 0 & 100 & \textbf{60.0} & 100 & 50 & 100 & \textbf{75.0} & 3 & -- & 2 & 1.67 & 28.15 & -- & 30.65 & 27.06 \\
\rowcolor{mecoorange}[\tabcolsep][\tabcolsep]\textbf{MECo-WAM} & 100 & 0 & 100 & \textbf{60.0} & 100 & 50 & 100 & \textbf{75.0} & 0 & -- & 1 & 0.83 & 25.69 & -- & 29.62 & \textbf{25.71} \\
\midrule
\multicolumn{17}{>{\columncolor{black!10}[\tabcolsep][\tabcolsep]}c}{\emph{Sort the three cubes into a line by size.}} \\
$\pi_0$          & 100 & 0 & 100 & 40.0 & 100 & 50 & 100 & 60.0 & 1 & -- & 1 & 1.50 & 22.88 & -- & 25.04 & \textbf{30.82} \\
Fast-WAM         & 100 & 100 & 100 & 60.0 & 100 & 100 & 100 & 75.0 & 0 & 4 & 0 & 1.33 & 26.03 & 57.94 & 25.23 & 38.49 \\
\rowcolor{mecoorange}[\tabcolsep][\tabcolsep]\textbf{MECo-WAM} & 100 & 100 & 100 & \textbf{70.0} & 100 & 100 & 100 & \textbf{80.0} & 0 & 2 & 3 & \textbf{1.00} & 26.35 & 28.15 & 55.47 & 31.96 \\
\iffalse
\midrule
\multicolumn{17}{@{}>{\columncolor{mecoredorange}[0pt][0pt]}c@{}}{\emph{Sort Toy.}} \\
$\pi_0$          & -- & -- & -- & -- & -- & -- & -- & -- & -- & -- & -- & -- & -- & -- & -- & -- \\
Fast-WAM         & -- & -- & -- & -- & -- & -- & -- & -- & -- & -- & -- & -- & -- & -- & -- & -- \\
\textbf{MECo-WAM} & -- & -- & -- & -- & -- & -- & -- & -- & -- & -- & -- & -- & -- & -- & -- & -- \\
\fi
\bottomrule
\end{tabular}%
}
\caption{Real-world tabletop results on two tasks (SR/PR/CR/CT: success rate/progress rate/correction count/completion time).}
\label{tab:real_world_main}
\end{table*}

\paragraph{Real-world evaluation.}
The real-world tabletop study evaluates \emph{Stack Cubes} and \emph{Sort Cubes by Size} on an ARX-R5 robotic arm, with representative rollouts shown in Figure~\ref{fig:real_robot}. Table~\ref{tab:real_world_main} compares $\pi_0$, Fast-WAM, and MECo-WAM using success rate, progress rate, correction count, and completion time. On \emph{Stack Cubes}, MECo-WAM matches Fast-WAM in SR/PR (60.0\%/75.0\%) but reduces corrections from 1.67 to 0.83 and shortens completion time from 27.06s to 25.71s, suggesting cleaner vertical alignment even when the final success rate is tied. On \emph{Sort Cubes by Size}, MECo-WAM improves Fast-WAM from 60.0\% to 70.0\% SR and from 75.0\% to 80.0\% PR, while reducing corrections from 1.33 to 1.00 and completion time from 38.49s to 31.96s. Averaged over the two tasks, MECo-WAM gains 5.0 SR points and 2.5 PR points over Fast-WAM, with about 39\% fewer corrections and 12\% shorter completion time, indicating that the learned 4D prior improves real-robot spatial grounding rather than only simulated benchmark scores.

\iffalse
\begin{table}[t]
\centering
\setlength{\tabcolsep}{8pt}
\begin{tabular}{@{}l|cccc@{}}
\toprule
\multicolumn{1}{@{}c|}{Method} & SR (\%) $\uparrow$ & PR (\%) $\uparrow$ & CR (\#) $\downarrow$ & CT (s) $\downarrow$ \\
\midrule
\multicolumn{5}{@{}>{\columncolor{black!10}[0pt][0pt]}c@{}}{\emph{Stack the three sponge cubes vertically (on the original tabletop).}} \\
$\pi_0$          & -- & -- & -- & -- \\
Fast-WAM         & -- & -- & -- & -- \\
\textbf{MECo-WAM} & -- & -- & -- & -- \\
\midrule
\multicolumn{5}{@{}>{\columncolor{black!10}[0pt][0pt]}c@{}}{\emph{Stack the three sponge cubes vertically (on one book).}} \\
$\pi_0$          & -- & -- & -- & -- \\
Fast-WAM         & -- & -- & -- & -- \\
\textbf{MECo-WAM} & -- & -- & -- & -- \\
\bottomrule
\end{tabular}
\caption{Height generalization on \emph{Stack Cubes}. The original tabletop condition matches the training setup, while the book conditions increase the working-surface height without changing the instruction. SR, PR, CR, and CT follow Table~\ref{tab:real_world_main}; the measured height offset for each book stack is recorded during evaluation.}
\label{tab:stack_height_generalization}
\end{table}

Table~\ref{tab:stack_height_generalization} isolates geometric generalization under vertical workspace shifts. Since the cube-stacking instruction is unchanged, performance changes primarily reflect whether the policy can adapt its spatial grounding, grasp height, and contact timing to an altered object height.
\fi

\paragraph{Representation probing.}

\begin{figure}[!t]
\centering
\includegraphics[width=1.00\columnwidth]{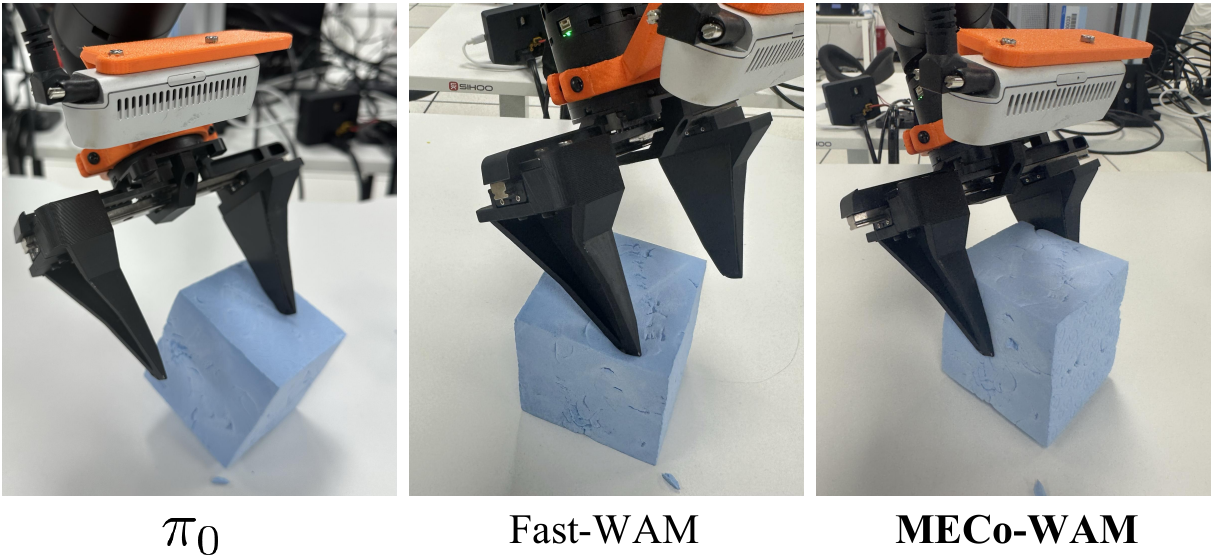}
\caption{3D position and pose sensitivity during real-robot grasping. MECo-WAM better preserves action-relevant cube position, grasp pose, and contact geometry.}
\label{fig:position}
\end{figure}

With the video-action backbone frozen, a matched DPT-style head \cite{ranftl2021dpt} on tokens from the final four layers shows sharper depth for MECo-WAM (Figure~\ref{fig:depthprobe}) \cite{li2026spatialforcing}. The grasp comparison in Figure~\ref{fig:position} further shows stronger sensitivity to action-relevant 3D position and pose. Both probes exclude auxiliary 4D tokens, indicating that geometry transfers into deployed video-action features.

\subsection{Ablation Studies}

\iffalse
\begin{table}[!htbp]
\centering
\scriptsize
\setlength{\tabcolsep}{2pt}
\resizebox{\columnwidth}{!}{%
\begin{tabular}{@{}lccccccc@{}}
\toprule
\multirow{2}{*}{Variant} & \multicolumn{3}{c}{Video} & Action & \multicolumn{3}{c}{Success Rate (\%) $\uparrow$} \\
\cmidrule(lr){2-4}\cmidrule(lr){5-5}\cmidrule(l){6-8}
& PSNR $\uparrow$ & SSIM $\uparrow$ & LPIPS $\downarrow$ & MSE ($\times 10$) $\downarrow$ & Clean & Random & Average \\
\midrule
Fast-WAM & 29.55 & 0.936 & 0.038 & 0.032 & 91.88 & 91.78 & 91.83 \\
+ 4D expert & 29.81 & 0.935 & 0.039 & 0.034 & 92.10 & 91.64 & 91.87 \\
+ decayed read & 30.06 & 0.939 & 0.038 & 0.026 & 92.36 & 91.92 & 92.14 \\
+ $\mathcal{L}_{\mathrm{geo}}^{\mathrm{act}}$ & 29.97 & 0.938 & 0.037 & 0.022 & 92.68 & 91.76 & 92.22 \\
+ $\mathcal{L}_{\mathrm{tem}}^{\mathrm{act}}$ & 30.42 & 0.940 & 0.039 & 0.019 & 92.42 & 92.08 & 92.25 \\
Full w/o aware & 30.31 & 0.942 & 0.038 & 0.017 & 92.90 & 91.86 & 92.38 \\
\rowcolor[gray]{0.92}
\textbf{Full MECo-WAM} & 30.72 & 0.942 & 0.037 & 0.013 & 93.26 & 91.98 & 92.62 \\
\bottomrule
\end{tabular}%
}
\caption{Ablation results on RoboTwin.}
\label{tab:ablation}
\end{table}
\fi

\begin{table}[!htbp]
\centering
\scriptsize
\setlength{\tabcolsep}{3pt}
\resizebox{\columnwidth}{!}{%
\begin{tabular}{lccccc}
\toprule
\multirow{2}{*}{\textbf{Variant}} & \multicolumn{3}{c}{\textbf{Video}} & \textbf{Action} & \textbf{Task} \\
\cmidrule(lr){2-4}\cmidrule(lr){5-5}\cmidrule(lr){6-6}
& \textbf{PSNR $\uparrow$} & \textbf{SSIM $\uparrow$} & \textbf{LPIPS $\downarrow$} & \textbf{MSE ($\times 10$) $\downarrow$} & \textbf{Avg SR $\uparrow$} \\
\midrule
Fast-WAM & 29.55 & 0.936 & 0.038 & 0.032 & 91.83 \\
+ 4D expert & 29.81 & 0.935 & 0.039 & 0.034 & 91.87 \\
+ decayed read & 30.06 & 0.939 & 0.038 & 0.026 & 92.14 \\
+ $\mathcal{L}_{\mathrm{geo}}^{\mathrm{act}}$ & 29.97 & 0.938 & 0.037 & 0.022 & 92.22 \\
+ $\mathcal{L}_{\mathrm{tem}}^{\mathrm{act}}$ & 30.42 & 0.940 & 0.039 & 0.019 & 92.25 \\
Full w/o aware & 30.31 & 0.942 & 0.038 & 0.017 & 92.38 \\
\rowcolor{mecoorange}[\tabcolsep][\tabcolsep]
\textbf{MECo-WAM} & 30.72 & 0.942 & 0.037 & 0.013 & 92.62 \\
\bottomrule
\end{tabular}%
}
\caption{Ablation results on RoboTwin.}
\label{tab:ablation}
\end{table}

Table~\ref{tab:ablation} isolates the proposed 4D co-training components under identical training and inference settings. We evaluate each variant from three aspects: video quality, action prediction, and task success. Adding only an isolated 4D expert yields a small task gain over Fast-WAM (91.83\% to 91.87\%) but does not improve action MSE (0.032 to 0.034), suggesting that geometric supervision is weak if it remains confined to an auxiliary branch. With decayed 4D read access, average success rises to 92.14\% and action MSE drops to 0.026, showing that temporary geometry-to-policy transfer is important. The spatial and temporal relation losses further reduce MSE to 0.022 and 0.019, respectively, and raise task success above 92.2\%. Full MECo-WAM obtains the best PSNR (30.72), the lowest action MSE (0.013), and the highest average success (92.62\%), a 0.79-point gain over Fast-WAM. The uniform-weight variant reaches 92.38\%, below the full model, indicating that the gain comes not only from additional 4D supervision but also from emphasizing action-relevant geometry.

\section{Conclusion}

We presented MECo-WAM, an inference-efficient WAM that injects 4D geometry only during training. A training-only 4D expert, decayed read-mask attention, and action-aware temporal geometric distillation transfer frozen VGGT priors into the deployed video-action representation, while all auxiliary geometry modules are removed at inference. Results on LIBERO, RoboTwin 2.0, and ARX-R5 real-world tasks, together with ablations, show that action-relevant 4D supervision improves geometric reasoning without increasing deployment cost.

\bibliography{aaai2027}

\end{document}